\documentclass[sigconf]{acmart}

\usepackage{algorithm}
\usepackage{algorithmic}
\usepackage{multirow}
\usepackage{enumitem}

\AtBeginDocument{%
  }

\copyrightyear{2025}
\acmYear{2025}
\setcopyright{cc}
\setcctype{by}
\acmConference[MM '25]{Proceedings of the 33rd ACM International Conference on Multimedia}{October 27--31, 2025}{Dublin, Ireland}
\acmBooktitle{Proceedings of the 33rd ACM International Conference on Multimedia (MM '25), October 27--31, 2025, Dublin, Ireland}\acmDOI{10.1145/3746027.3755532}
\acmISBN{979-8-4007-2035-2/2025/10}

\begin{document}

\title[InterAnimate]{InterAnimate: Taming Region-Aware Diffusion Model for \\ Realistic Human Interaction Animation}

\author{Yukang Lin}
\authornote{Both authors contributed equally to this research.}
\orcid{0009-0001-2469-5690}
\affiliation{%
  \institution{Tsinghua University}
  \city{Shenzhen}
  \country{China}
}
\email{linyk23@mails.tsinghua.edu.cn}

\author{Yan Hong}
\orcid{0000-0001-6401-0812}
\authornotemark[1]
\affiliation{%
  \institution{Ant Group}
  \city{Hangzhou}
  \country{China}
}
\email{ruoning.hy@antgroup.com}

\author{Zunnan Xu}
\orcid{0000-0001-5586-4971}
\affiliation{%
  \institution{Tsinghua University}
  \city{Shenzhen}
  \country{China}}
\email{xzn23@mails.tsinghua.edu.cn}

\author{Xindi Li}
\orcid{0009-0002-2890-6958}
\affiliation{%
  \institution{Zhejiang University}
  \city{Hangzhou}
  \country{China}}
\email{xindili@zju.edu.cn}

\author{Chao Xu}
\orcid{0009-0004-8661-4579}
\affiliation{%
  \institution{Ant Group}
  \city{Hangzhou}
  \country{China}
}
\email{yanyue.xc@antgroup.com}

\author{Chuanbiao Song}
\orcid{0009-0002-7466-9383}
\affiliation{%
  \institution{Ant Group}
  \city{Hangzhou}
  \country{China}
}
\email{songchuanbiao.scb@antgroup.com}

\author{Ronghui Li}
\orcid{0000-0001-7703-9315}
\affiliation{%
  \institution{Tsinghua University}
  \city{Shenzhen}
  \country{China}}
\email{lrh22@mails.tsinghua.edu.cn}

\author{Haoxing Chen}
\orcid{0000-0001-6637-8741}
\affiliation{%
  \institution{Ant Group}
  \city{Hangzhou}
  \country{China}
}
\email{chenhaoxing.chx@antgroup.com}

\author{Jun Lan}
\orcid{0000-0003-0921-0613}
\affiliation{%
  \institution{Ant Group}
  \city{Hangzhou}
  \country{China}
}
\email{yelan.lj@antgroup.com}

\author{Huijia Zhu}
\orcid{0009-0008-5784-7225}
\authornote{Both authors are the corresponding author.}
\affiliation{%
  \institution{Ant Group}
  \city{Hangzhou}
  \country{China}
}
\email{huijia.zhj@antgroup.com}

\author{Weiqiang Wang}
\orcid{0000-0002-6159-619X}
\affiliation{%
  \institution{Ant Group}
  \city{Hangzhou}
  \country{China}
}
\email{weiqiang.wwq@antgroup.com}

\author{Jianfu Zhang}
\orcid{0000-0002-2673-5860}
\authornotemark[2]
\affiliation{%
 \institution{Shanghai Jiao Tong University}
  \city{Shanghai}
  \country{China}}
\email{c.sis@sjtu.edu.cn}

\author{Xiu Li}
\orcid{0000-0003-0403-1923}
\authornotemark[2]
\affiliation{%
  \institution{Tsinghua University}
  \city{Shenzhen}
  \country{China}}
\email{li.xiu@sz.tsinghua.edu.cn}

\renewcommand{\shortauthors}{Yukang Lin et al.}

\begin{abstract}
Recent video generation research has focused heavily on isolated actions, leaving interactive motions—such as hand-face interactions—largely unexamined. These interactions are essential for emerging biometric authentication systems, which rely on interactive motion-based anti-spoofing approaches. From a security perspective, there is a growing need for large-scale, high-quality interactive videos to train and strengthen authentication models. In this work, we introduce a novel paradigm for animating realistic hand-face interactions. Our approach simultaneously learns spatio-temporal contact dynamics and biomechanically plausible deformation effects, enabling natural interactions where hand movements induce anatomically accurate facial deformations while maintaining collision-free contact. To facilitate this research, we present \textbf{InterHF}, a large-scale hand-face interaction dataset featuring 18 interaction patterns and 90,000 annotated videos. Additionally, we propose \textbf{InterAnimate}, a region-aware diffusion model designed specifically for interaction animation. InterAnimate leverages learnable spatial and temporal latents to effectively capture dynamic interaction priors and integrates a region-aware interaction mechanism that injects these priors into the denoising process. To the best of our knowledge, this work represents the first large-scale effort to systematically study human hand-face interactions. Qualitative and quantitative results show InterAnimate produces highly realistic animations, setting a new benchmark. Code and data will be made public to advance research.
\end{abstract}






\maketitle

\section{Introduction}

\begin{figure}
    \centering
    \includegraphics[width=0.9\linewidth]{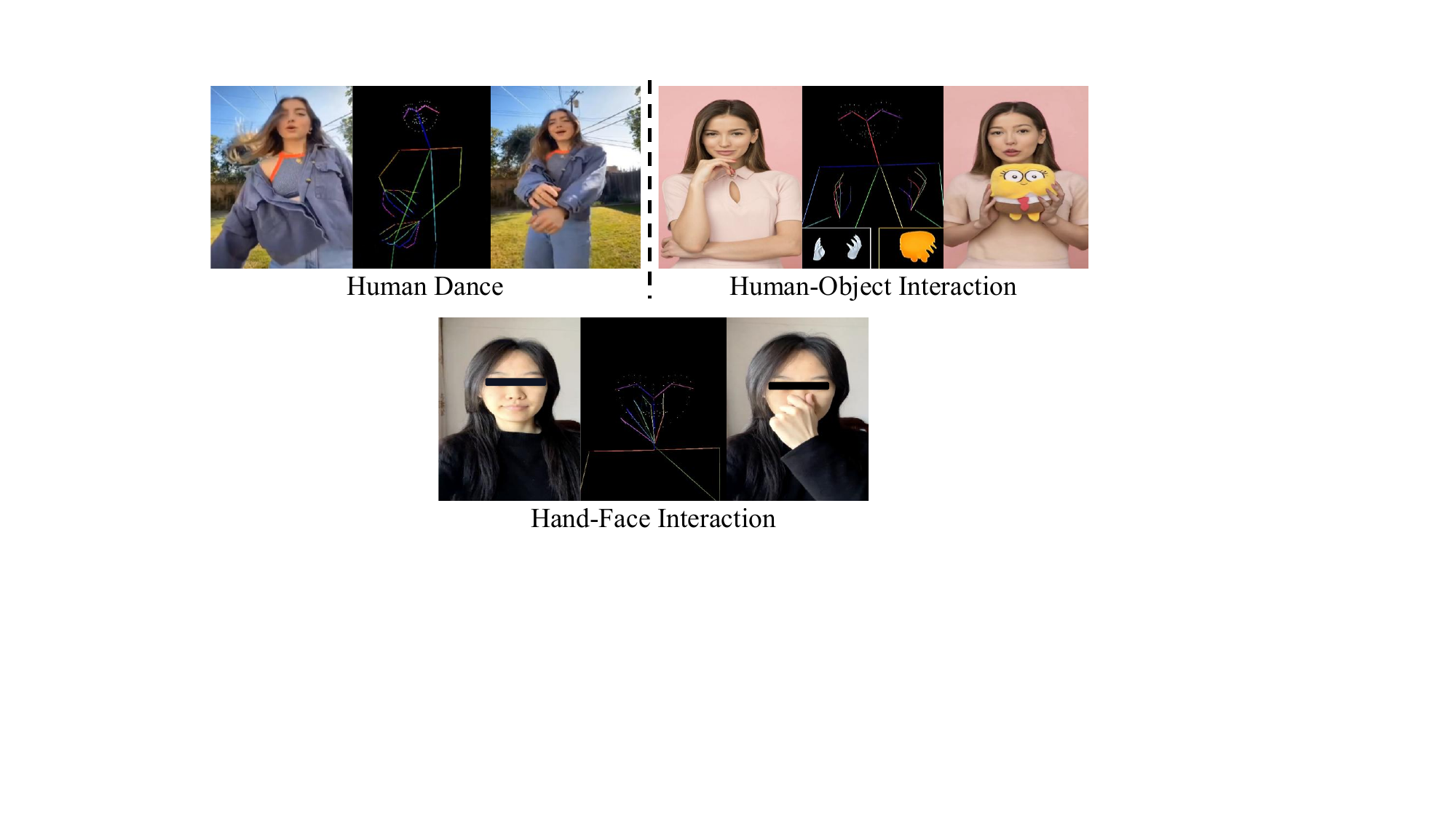}
     \vspace{-5pt}
    \caption{Comparison of three types of human animation video generation: human dance, human-object interaction, and hand-face interaction.}
   \vspace{-10pt}
    \label{fig:motivation}
\end{figure}

% Recently, the rapid advancements in video generation have significantly propelled research in the field of human animation. This progress has enabled the creation of increasingly realistic and interesting human videos, with applications spanning virtual reality, gaming, movie production, and beyond. In this paper, we focus on pose-driven human animation conditioned on a reference image.
Recent advances in video generation have significantly advanced human animation research, enabling increasingly realistic videos with broad applications in virtual reality, gaming, and entertainment. Popular diffusion models, such as the series of image generation ~\cite{rombach2021highresolution} and the video generation model~\cite{blattmann2023stable} of Stable Diffusion, have shown their effectiveness in generating high-quality images and videos. Researchers are increasingly investigating human image-to-video tasks by utilizing the architecture of these models. As a pioneer in human dance generation, Disco~\cite{Wang_2024_CVPR} selected U-Net as its backbone and developed two ControlNets to manage the background and human pose. Following this, AnimateAnyone~\cite{hu2024animate} and MagicAnimate~\cite{xu2023magicanimate} improved the fluidity of the animations by integrating motion modules into the U-Net. More recently, to fully leverage the video priors learned from large-scale video datasets, MimicMotion~\cite{mimicmotion2024}, Dispose~\cite{li2024dispose}, and StableAnimator~\cite{tu2024stableanimator} adopted the Stable Video Diffusion as their backbone. However, most existing work focuses on isolated actions, such as dancing, and largely overlooks interactive motions—particularly those involving hands and faces. These interactions are crucial for next-generation biometric authentication systems~\cite{anthony2021review,muhammad2024saliency}, which rely on dynamic, interaction-based anti-spoofing measures (e.g., “rub your chin,” “cover your left eye”) that demand high-fidelity interactive motion generation. Despite their importance, such interactions remain underexplored in the current research landscape.

% Almost all human animation work focuses on dance task, while paying limited attention to interaction, particularly between the expressive body parts like hands and faces. 

% In this paper, we introduce a novel animation concept centered on interaction between the hand and face. This task remains underexplored, leaving a significant gap in the development of realistic and engaging human animations.
% Unlike prior work that primarily emphasizes isolated movements or human-object interactions~\cite{xu2024anchorcrafter}, our study focuses on the complex patterns of hand-face interaction and the deformations that arise from physical contact. Our objective is to achieve highly realistic and natural interaction animations that accurately capture the dynamics of hand-face contact.

To address this gap, we introduce a novel paradigm for animating realistic hand-face interactions. Unlike previous work that focuses on isolated movements or human-object interactions~\cite{xu2024anchorcrafter}, as shown in Fig. \ref{fig:motivation}, our approach focuses on the complex dynamics and deformation effects that emerge from physical contact between hands and faces. Our goal is to achieve highly realistic, natural interactive animations that faithfully capture the nuances of hand-face contact.

To this end, we collect a comprehensive dataset, \textbf{InterHF}, a new large-scale dataset specifically designed to advance hand-face interaction modeling. \textbf{InterHF} captures 18 distinct interaction patterns across four major categories: pinching, stroking, poking, and swiping gestures. Each category is further divided into detailed sub-classes, providing comprehensive coverage of a diverse range of motions and contact scenarios. The dataset consists of 90,000 annotated video clips, totaling 75 hours of high-quality footage. This extensive resource not only enables detailed analysis of the fine-grained dynamics of hand-face interactions but also serves as a critical foundation for synthesizing these complex motions with unprecedented realism and precision.

% To this end, we collect a comprehensive dataset, \textbf{InterHF}, which comprises 18 distinct interaction patterns and a total of 90,000 high-quality videos. 
% This dataset serves as a foundational resource for modeling and synthesizing hand-face interactions with unprecedented detail and realism.

\begin{figure}
 \centering
  \includegraphics[width=0.8\linewidth]{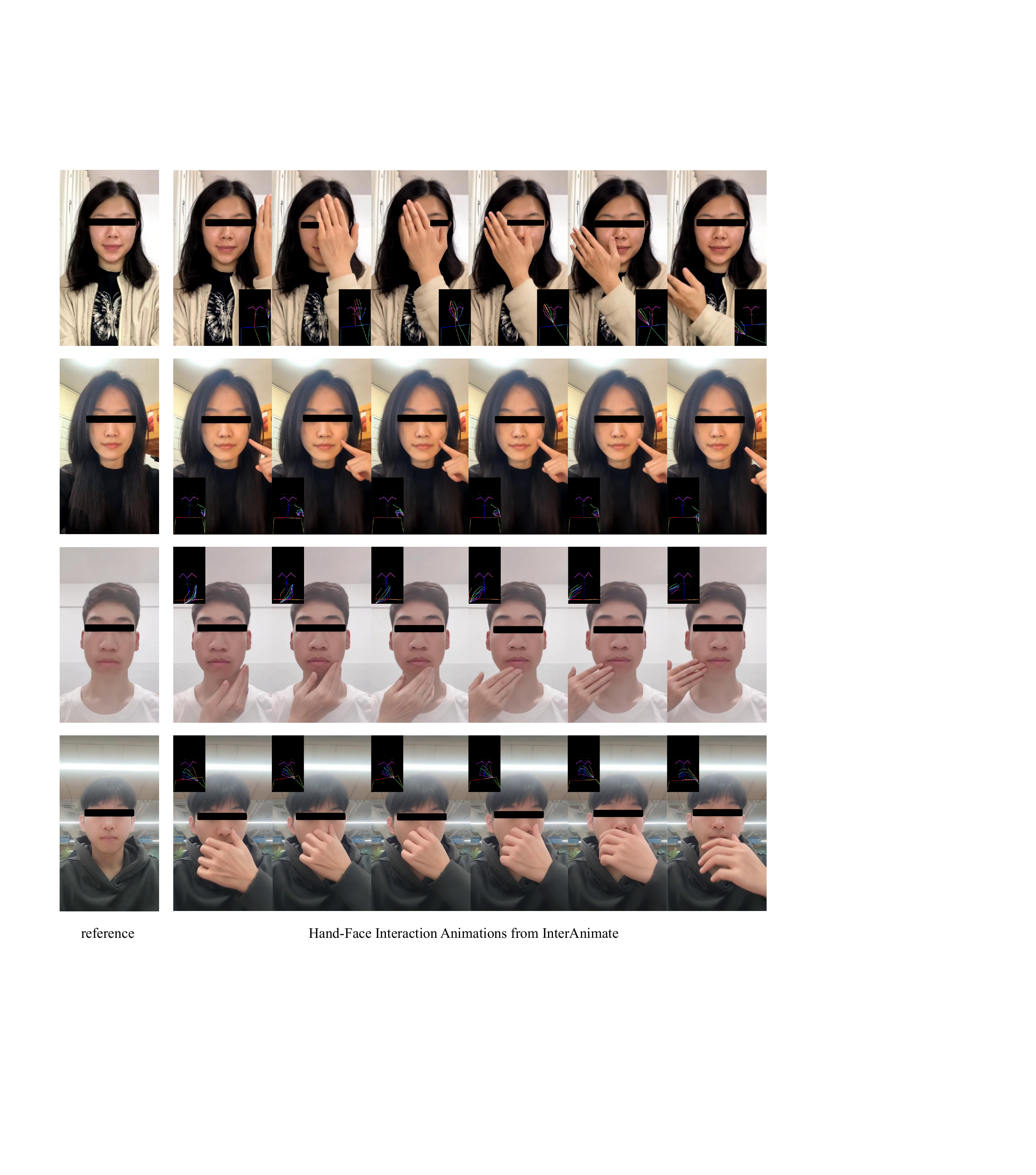}
  \vspace{-5pt}
  \caption{The highly realistic interaction animations generated from InterAnimate.}
  \label{fig:teaser}
  \vspace{-4mm}
\end{figure}

Building on \textbf{InterHF} dataset, we propose \textbf{InterAnimate}, a novel region-aware diffusion model specifically designed for interaction animation. 
InterAnimate introduces a region-aware interaction system that significantly enhances its ability to model the intricate dynamics of hand-face contact. Central to this system are learnable interaction latents, which include both spatial and temporal components designed to capture rich interaction priors. To seamlessly integrate these priors into the denoising process, we develop a specialized region attention block that operates in three key stages: soft quantization, cross-attention, and masking.
To maintain identity consistency throughout the hand-face interactions, we implement an ID Preserver, which utilizes facial embeddings derived from ArcFace~\cite{deng2019arcface}. Additionally, we incorporate an orthogonality loss to encourage independence among the interaction latents, thereby increasing their capacity to learn a diverse array of interaction patterns. These innovations collectively enable \textbf{InterAnimate} to produce animations that not only achieve realistic and accurate contact, but also maintain smooth, coherent motion over time, as illustrated in Fig. \ref{fig:teaser}.
% A key innovation of InterAnimate is the introduction of a region-aware interaction system, which enhances its ability to model the intricate contact patterns between hands and faces. This system is centered around learnable interaction latents, which consist of spatial and temporal latents that capture rich priors of interaction. To effectively inject these priors into the denoising process, we design a region attention block, which employs three steps: soft quantization, cross-attention, and masking. Furthermore, we introduce an ID Preserver that leverages face embeddings extracted via ArcFace to ensure identity consistency during hand-face interactions. Besides, we introduce the orthogonality loss to encourages independence among the interaction latents, enhancing the latents' capacity to learn diverse interaction patterns. All these designs enable InterAnimate to generate animations that not only exhibit realistic contact but also preserve the natural flow and coherence of movement over time.

% We evaluate \textbf{InterAnimate} on the \textbf{InterHF} test set. Through quantitative and qualitative analysis, we demonstrate the superiority of \textbf{InterAnimate} when compared to state-of-the-art methods.

Through comprehensive qualitative and quantitative evaluation on the \textbf{InterHF} test set, we demonstrate that \textbf{InterAnimate} achieves superior performance compared to state-of-the-art methods, delivering animations that set a new benchmark in the field. 
In summary, our contributions can be summarized as follows:
\begin{itemize}[leftmargin=*,noitemsep,nolistsep]
% \begin{itemize}
    % \setlength{\itemindent}{-1.5em}
    \item \textbf{InterHF}, the first dataset dedicated to modeling hand-face interactions with a wide range of patterns, addressing a critical gap in existing resources.
    \item \textbf{InterAnimate}, a region-aware diffusion model tailored for interaction animation, featuring innovations such as a region-aware attention system and ID Preserver.
    \item A pioneering exploration of hand-face interactions in video generation, laying the groundwork for future research in complex human interaction animation.
\end{itemize}
\section{Related Work}
\subsection{Diffusion for Video Generation}  
% Recent advances in video generation have extended diffusion models from static image synthesis to dynamic sequence modeling. 
Early video diffusion approaches~\cite{singer2022make} pioneered temporal modeling through 3D U-Nets and spatio-temporal attention. Subsequent works like Video Diffusion Models~\cite{he2022lvdm} introduced cascaded super-resolution pipelines for quality enhancement, while Tune-A-Video~\cite{wu2023tune} enabled one-shot video adaptation through parameter-efficient fine-tuning. The emergence of Stable Video Diffusion~\cite{blattmann2023svd} established new benchmarks in temporal consistency via large-scale video dataset training and frame-wise latent alignment. 
% Notably, human animation research has leveraged these advances through architectures like Disco~\cite{Wang_2024_CVPR} and AnimateAnyone~\cite{hu2024animate}, which integrate ControlNets and motion modules into U-Net backbones for pose-guided synthesis. 
% Recent paradigms such as StableAnimator~\cite{tu2024stableanimator} further exploit video diffusion priors by adopting Stable Video Diffusion as their foundation model.  
While models like HunyuanVideo~\cite{kong2024hunyuanvideo} and SORA~\cite{lin2024open} demonstrate the potential of large-scale training for generating long-form coherent videos, current methods largely focus on full-body dance~\cite{li2024dispose} or rigid object manipulation, neglecting the dynamics of expressive part interactions such as hand-face coordination. In this work, we bridge this gap by developing a region-aware interaction diffusion framework that combines global pose guidance with localized attention mechanisms, enabling synchronized control of both macro-body movements and micro-level facial-hand interactions while maintaining temporal coherence across video sequences.

\subsection{Pose-guided Human Image Animation}
Pose-guided human animation leverages pose information to generate realistic motion, with recent advances propelled by diffusion models~\cite{text2videozero,tuneavideo,rerender}, particularly those built on the Stable Diffusion framework~\cite{rombach2021highresolution, blattmann2023stable}, which refine motion dynamics iteratively. Poses are typically encoded as keypoints or skeletons and integrated via ControlNet~\cite{zhang2023controlnet} or ReferenceNet~\cite{hu2024animateanyone}. Pioneering works like Disco~\cite{Wang_2024_CVPR} established the paradigm of combining U-Net architectures with dedicated ControlNets for background and pose management, while subsequent innovations such as AnimateAnyone~\cite{hu2024animate} and MagicAnimate~\cite{xu2023magicanimate} enhanced motion fluidity through specialized motion modules.
%
% A critical component of this progress lies in the evolution of pose estimation techniques. OpenPose~\cite{openpose} pioneered real-time multi-person keypoint detection, providing foundational skeletal data for animation tasks. Subsequent developments, such as DWpose~\cite{dwpose}, further enhanced accuracy by addressing challenges in occluded or ambiguous poses through frequency-aware decomposition. Complementing these efforts, DensePose~\cite{densepose} introduced dense correspondences between 2D images and 3D body surfaces, enabling pixel-level control over pose variations. For parametric modeling of human figures, the SMPL framework~\cite{SMPL} has become a standard tool for representing pose and shape parameters. Its vertex-based design supports both coarse body motion synthesis and fine-grained deformations, as demonstrated in recent animation studies. Researchers have further combined SMPL priors with diffusion models~\cite{shao2024human4dit} to enhance temporal coherence in video generation while preserving realism.
%
Existing methods aim to preserve identity features from reference images and synthesize pose-driven motions. While state-of-the-art approaches~\cite{tu2024stableanimator} employ spatial-temporal attention mechanisms to separate appearance attributes from motion dynamics, recent trends~\cite{li2024dispose,mimicmotion2024} show increasing adoption of video diffusion architectures like Stable Video Diffusion~\cite{blattmann2023svd} to leverage large-scale video priors. However, these works predominantly focus on full-body dance generation, neglecting nuanced interactions between expressive body parts such as hands and facial features. Building upon these foundations, our method introduces a region-aware interaction system that explicitly models hand-face interactions in complex animation scenarios.

\subsection{Human Animation with Interaction}
Human animation with interaction aims to depict realistic relationships between humans and objects. While current methods primarily concentrate on human-object interaction (HOI) generation~\cite{xu2024anchorcrafter} or object swapping in hand-centric videos, they lack sufficient degrees of freedom for expressive body part interactions. This limitation becomes particularly evident in \textit{hand-face interaction} scenarios that require simultaneous management of facial deformation, partial occlusion, and identity preservation during contact.
Recent advances in HOI generation have facilitated the synthesis of rigid object manipulation~\cite{li2023object,peng2023hoi,xu2023interdiff}, but remain inadequate for modeling soft tissue dynamics. Traditional 3D object representations~\cite{ghosh2023imos} fail to capture non-rigid facial deformations, while spatial editing methods like HOI-Swap~\cite{xue2024hoi} neglect topological changes during hand-face contact. Three fundamental limitations persist across existing approaches: (i) absence of physics-inspired facial distortion models during finger contact, (ii) inability to maintain identity consistency under dynamic hand-induced occlusions, and (iii) restricted focus on isolated interactions rather than coordinated full-body motions.
To address these challenges, we introduce the InterHF dataset, which contains 90,000 high-quality interaction videos. Our approach enables the generation of natural hand-face interaction, establishing new benchmarks for complex animation with interaction.

% and coordinated body language

\begin{figure*}[htbp]
  \centering
  \includegraphics[width=0.95\textwidth]{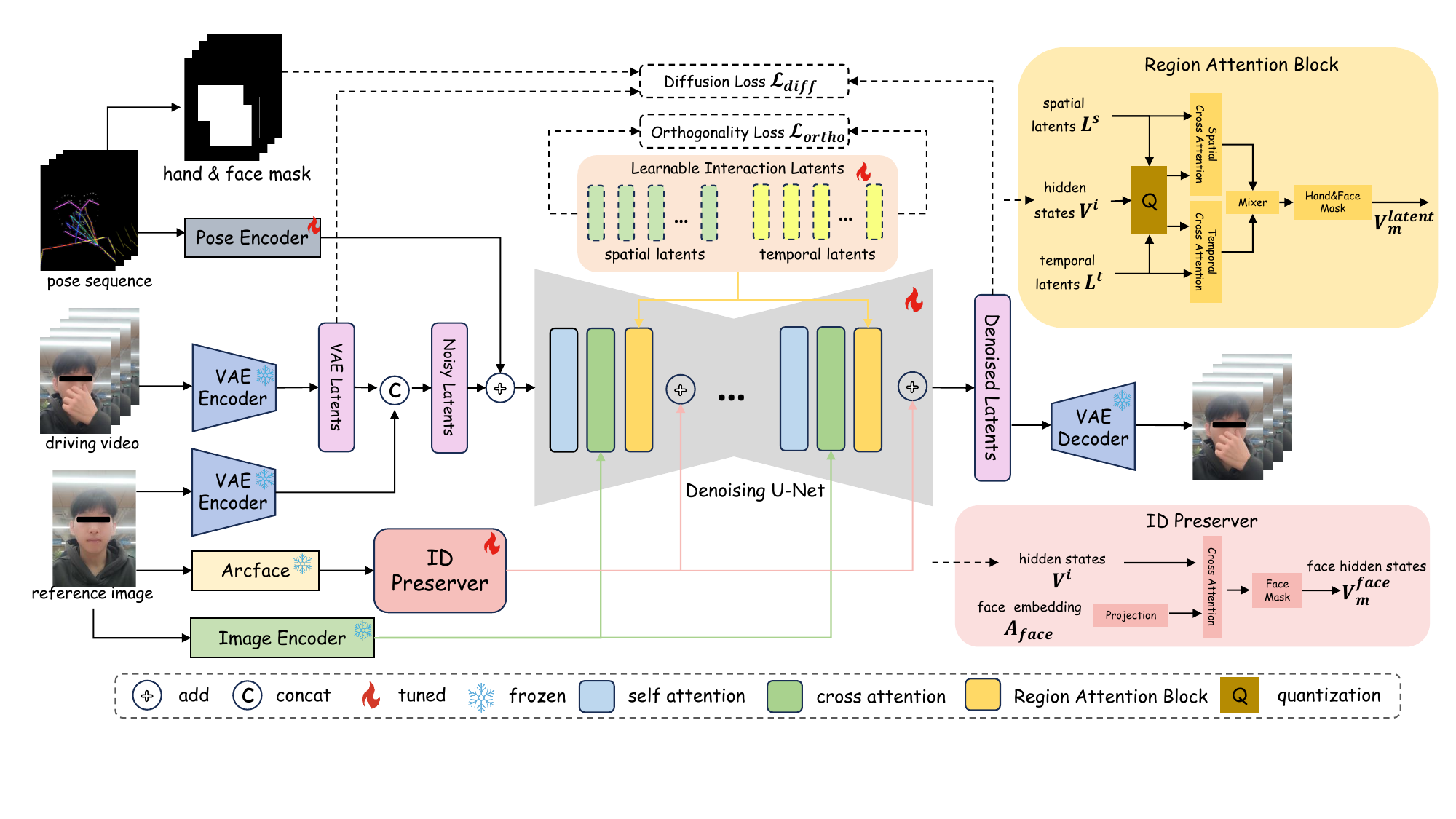}
  \vspace{-5pt}
   \caption{The framework of InterAnimate.}
  \label{fig:pipline}
  \vspace{-0.2cm}
\end{figure*}

\section{InterHF Dataset Collection}
To advance research in hand-face interaction synthesis, we introduce \textbf{InterHF}, a large-scale, multi-modal dataset offering several distinct advantages. InterHF captures a diverse range of environmental conditions, including varied lighting setups and multiple backgrounds. The dataset also incorporates significant photometric diversity—such as specular highlights and shadows—alongside geometric variability in hand and face shapes from a balanced demographic sample. Moreover, it reflects natural behavioral patterns, including variations in interaction speed, contact force, and motion dynamics. This makes InterHF uniquely valuable for developing robust, real-world interaction synthesis models, filling a critical gap in human-centric animation research.
Next, we detail the data collection protocol, interaction taxonomy, and dataset statistics.

\subsection{Data Collection Protocol}
The \textbf{InterHF} dataset was constructed through a meticulously planned data collection procedure, encompassing a total of 5000 participants. These individuals were evenly distributed by gender—3000 male and 2000 female—and categorized into ten distinct age groups spanning from infancy to advanced age (0–10, 11–20, ..., 91–100), ensuring balanced demographic representation across the entire population.

Data collection took place under eight carefully controlled environmental scenarios designed to capture a broad spectrum of real-world conditions. These scenarios incorporated variations in lighting direction (front-lit, back-lit), lighting sources (natural daylight, artificial point lighting), and recording locations (indoor laboratories, outdoor urban or park settings). To ensure the acquisition of diverse and high-quality data streams, multiple synchronized devices were employed. Specifically, the setup included ten smartphone models (e.g., iPhone series, Huawei devices) and ten webcam models (e.g., MacBooks, Samsung PCs), enabling robust multimodal recording. This carefully crafted setup provided not only a wide range of hand-face interaction data but also comprehensive ground-truth annotations that facilitate detailed analysis and modeling.

\subsection{Interaction Taxonomy}
The \textbf{InterHF} dataset establishes a robust and methodical taxonomy for hand-face interactions, categorized into four primary types: \textit{\textbf{pinching gestures}}, \textit{\textbf{stroking gestures}}, \textit{\textbf{poking gestures}}, and \textit{\textbf{swiping gestures}}. Within these overarching categories, 18 well-defined sub-classes capture a diverse range of movement patterns and facial contact dynamics. Each class was meticulously designed to encompass key variations in motor control and facial interaction, providing a foundation for detailed analysis and modeling.

% facial interaction parameters

The \textit{\textbf{pinching gestures}} category includes two distinct sub-classes: left-hand nose pinch (LH-NP) and right-hand nose pinch (RH-NP). In contrast, \textit{\textbf{stroking gestures}} span eight sub-classes, covering four specific facial regions—eyebrow, forehead, chin, and ear—with each action performed by both the left and right hands. This results in sub-classes such as LH-EB, LH-FH, LH-CH, LH-ER, and their respective right-hand counterparts (RH-EB, RH-FH, RH-CH, RH-ER). The \textit{\textbf{poking gestures}} category consists of six sub-classes that systematically combine cheek laterality (left/right) and digit configurations (one, two, or three fingers). For example, single-finger pokes are represented as LC-SF and RC-SF, while two- and three-finger variants are denoted as LC-TF, RC-TF, LC-TH, and RC-TH. All poking interactions maintain a standardized approach angle of 30°–45° relative to the facial plane to ensure consistency in deformation patterns. Finally, \textit{\textbf{swiping gestures}} comprise two main sub-classes: left-to-right (LR-FS) and right-to-left (RL-FS) palmar trajectories. These gestures were carefully delineated to reflect natural directional movements of the palm. By organizing these gestures into a structured hierarchy, the dataset facilitates detailed investigation into both the kinematic control of hand movements and their corresponding facial deformation effects. This taxonomy, rigorously defined and uniformly applied, provides a consistent framework for benchmarking and further research in the field.

\subsection{Dataset Statistics}
The \textbf{InterHF} dataset consists of 90,000 meticulously annotated video clips, evenly divided among four primary interaction categories, each of which contains eighteen sub-classes with 5,000 samples per sub-class. This translates to approximately 75 hours of diverse, high-quality footage. The dataset is split into distinct subsets for model training and evaluation, with 90\% of the clips allocated for training and the remaining 10\% reserved for testing.

The dataset’s extensive scale and detailed annotations, combined with its photometric and geometric diversity, offer a significant improvement over existing datasets such as TikTok~\cite{jafarian2021learning} and Ted-talk ~\cite{siarohin2021motion} dataset. By providing a wide range of natural behavioral variations and robust ground-truth data, InterHF sets a new benchmark for real-world interaction synthesis research.
\section{Methodology}
\textbf{InterAnimate} is composed of two core modules that jointly enable the generation of realistic hand-face interaction animations: the \textbf{Region-aware Interaction System} and the \textbf{ID Preserver}. The Region-aware Interaction System introduces learnable interaction latents and a region attention mechanism to effectively model complex spatio-temporal contact patterns. Meanwhile, the ID Preserver ensures facial identity consistency by leveraging ArcFace embeddings and integrating them into the animation process. Together, these modules allow InterAnimate to produce animations that maintain both natural motion coherence and visual fidelity.
\subsection{Region-aware Interaction System}
The Region-aware Interaction System forms the backbone of InterAnimate’s animation process. 
% In order to generate realistic human interaction animation, we design a region-aware interaction system. 
% Specifically, we first establish two sets of Learnable Interaction Latents, including spatial latents and temporal latents. We anticipate that after training, these learnable vector can be used to represent different interaction modes. Second, we design the Region Attention Block, which is used to inject the interaction patterns expressed in the vector groups into the denoising process of the video diffusion model.
It introduces two sets of Learnable Interaction Latents, including spatial latents and temporal latents, to capture diverse interaction priors. These latents learn from training samples to represent distinct hand-face interaction patterns over time. To integrate these interaction priors into the animation process, we propose the Region Attention Block to refine latent features, enrich spatial and temporal representations, and focus computation on interactive regions, enabling anatomically accurate and visually coherent animation generation.

\textbf{Learnable Interaction Latents.} 
There are lots of works that utilize learnable parameters to extract specific features~\cite{li2023blip} or to learn motion patterns ~\cite{zhang2023generating, xu2024mambatalk, lin2024cyberhost}. Inspired by this, we construct two sets of learnable vectors, dubbed Learnable Interaction Latents in our paper, to learn the interaction priors from the training samples. Specifically, spatial latents $\mathbf{L}^\mathrm{s~}\in\mathbb{R}^{n\times d}$ are used to learn spatial features such as the topology and texture of hands and faces, as well as the the appearance of the contact area. The temporal latents $\mathbf{L}^\mathrm{t~}\in\mathbb{R}^{m\times d}$ are used to learn temporal features such as interactive action patterns and contact deformation processes. The $n$ and $m$ denote the number of learnable vectors and $d$ denotes the vector dimension. During training, the learnable interaction latents can continuously learn from interaction samples and self-update, resembling a process of acquiring knowledge about interactions.

\textbf{Region Attention Block.} To inject the interaction priors stored in the learnable interaction latents into the denoising process of the diffusion model, we designed a region attention block, which is illustrated in Fig. \ref{fig:pipline}. Injection can be divided into \textbf{three steps}: soft quantization, cross attention, and masking. 

\begin{algorithm}[H]
    \caption{Quantize Hidden States with Soft Nearest Neighbor}
    \renewcommand{\algorithmicrequire}{ \textbf{Input:}}
    \renewcommand{\algorithmicensure}{ \textbf{Output:}} 
    \label{alg: quantize}
    \begin{algorithmic}[1]

        \REQUIRE % \textbf{Input:} 
        % $\text{batch size } b$, $\text{frame number } f$, 
        % $\text{latent number } n$, 
        $ \text{hidden states } \mathbf{V}^\mathrm{i} \in \mathbb{R}^{b \times f \times h \times w \times c}$, $\text{interaction latents } \mathbf{L} \in \mathbb{R}^{n \times d}$, $ \text{temperature } \tau \in \mathbb{R}$
        
        \ENSURE %\textbf{Output:} 
        $\text{quantized hidden states } \mathbf{\bar{V}} $
        
        \STATE $\mathbf{V}_\mathrm{flatten} = \mathbf{V}^\mathrm{i}.\text{view} (-1, c))$      \COMMENT{reshape to (b*f*h*w, c)}
        
        \STATE distance $D = \sum_{c}  \mathbf{V}_\mathrm{flatten}^2 + \sum_{d} \mathbf{L}^2 - 2 \times \text{matmul}( \mathbf{V}_\mathrm{flatten}, \mathbf{L}^\top)$
        \STATE $w = \text{softmax}(-D / \tau, dim=1)$
        
        \STATE $out = \text{matmul}(w, \mathbf{L})$ \COMMENT{weighted sum}
        \STATE $\mathbf{\bar{V}} = out.\text{view} (b, f, h, w, c))$
        
        \STATE \textbf{Return:} $\mathbf{\bar{V}}$
    \end{algorithmic}
\end{algorithm}

% Firstly, we quantize the hidden states $\mathbf{F}^\mathrm{in}$, the input of the region attention block, to obtain $\mathbf{F}^\mathrm{spa}_\mathrm{quantized}$ and $\mathbf{F}^\mathrm{temp}_\mathrm{quantized}$. The quantization uses soft nearest neighbor instead of hard indexing used in codebook. The soft quantization is detailed in Algorithm.\ref{alg: quantize}.
Firstly, we perform a quantization step on the input hidden states, denoted as $\mathbf{V}^\mathrm{i}$, which results in the generation of spatially quantized hidden states $\mathbf{\bar{V}}^\mathrm{s}$ and temporally quantized hidden states $\mathbf{\bar{V}}^\mathrm{t}$. Unlike the traditional hard indexing approach found in codebooks, this quantization method employs a soft nearest neighbor mechanism, as detailed in Algorithm.\ref{alg: quantize}. The proposed soft quantization approach provides several key advantages: (i) By enabling each hidden state to draw on multiple interaction latents, it enriches the representation of interaction information, thereby improving adaptability to complex patterns and enhancing generalization. (ii) it promotes smooth transitions between states, avoiding abrupt shifts and making it easier to identify latent movement patterns. (iii) Its differentiable nature ensures a more stable and efficient training process.

% The proposed soft quantization approach offers several advantages: (i) By allowing each hidden state to be quantized based on multiple interaction latents, our method can integrate rich interaction information and adapt to complex patterns, leading to improved generalization capabilities. (ii) This approach facilitates smooth transitions during state changes, as opposed to abrupt shifts to a specific mode, thereby aiding in the discovery of latent movement patterns. (iii) The differentiable nature of the soft quantization process contributes to a more stable and efficient training regimen.

% Next, we will obtain the enhanced spatial hidden states $\mathbf{F}^\mathrm{spa}$ and enhanced temporal hidden states $\mathbf{F}^\mathrm{temp}$ by conducting spatial and temporal cross attention. Then, a Mixer is used to fuse $\mathbf{F}^\mathrm{spa}$ and $\mathbf{F}^\mathrm{temp}$ to get $\mathbf{F}^\mathrm{latent}$. The process can be formulated as
Following the soft quantization process, we apply spatial and temporal cross-attention mechanisms to derive the enhanced spatial hidden states $\mathbf{\hat{V}}^{\mathrm{s}}$ and enhanced temporal hidden states $\mathbf{\hat{V}}^{\mathrm{t}}$. These refined representations are then combined using a Mixer, which integrates $\mathbf{\hat{V}}^{\mathrm{s}}$ and $\mathbf{\hat{V}}^{\mathrm{t}}$ to produce the fused latent states $\mathbf{V}^\mathrm{latent}$. The complete process can be mathematically expressed as follows:

\begin{equation}
\begin{aligned}
\mathbf{\hat{V}}^{\mathrm{s}}&=\mathrm{SpatialCrossAttn}(\mathbf{\bar{V}}^\mathrm{s},\mathbf{L}^\mathrm{s},\mathbf{L}^\mathrm{s}),
% \\ 
% &=\mathrm{softmax}\left(\frac{\mathbf{QK}_{\mathrm{spa}}^T}{\sqrt{d}}\right)\cdot\mathbf{V}_{\mathrm{spa}},
\end{aligned}
\end{equation}

\begin{equation}
\begin{aligned}
\mathbf{\hat{V}}^{\mathrm{t}}&=\mathrm{TemporalCrossAttn}(\mathbf{\bar{V}}^\mathrm{t},\mathbf{L}^\mathrm{t},\mathbf{L}^\mathrm{t}),
% \\ &=\mathrm{softmax}\left(\frac{\mathbf{QK}_{\mathrm{temp}}^T}{\sqrt{d}}\right)\cdot\mathbf{V}_{\mathrm{temp}},
\end{aligned}
\end{equation} 

\begin{equation}
\begin{aligned}
\mathbf{V}^\mathrm{latent}=\mathrm{Mixer}(\mathbf{\hat{V}}^{\mathrm{s}}, \mathbf{\hat{V}}^{\mathrm{t}}) = \alpha \cdot \mathbf{\hat{V}}^{\mathrm{s}} + (1 - \alpha) \cdot \mathbf{\hat{V}}^{\mathrm{t}},
\end{aligned}
\end{equation} 
% where $\mathbf{Q}$, $\mathbf{K}$ and $\mathbf{V}$ are the query, key, and value, respectively. The hidden state serves as the query, while the interaction latents act as the key and value. 
$\mathrm{SpatialCrossAttn}$ and $\mathrm{TemporalCrossAttn}$ operations all using the standard cross-attention mechanism $\mathrm{CrossAttn}$:
\[
\mathrm{CrossAttn}(\mathbf{Q}, \mathbf{K}, \mathbf{V}) = \mathrm{softmax}\left(\frac{\mathbf{Q}\mathbf{K}^\mathrm{T}}{\sqrt{d}}\right) \cdot \mathbf{V},
\]
where $\mathbf{Q}$, $\mathbf{K}$, and $\mathbf{V}$ denote the query, key, and value, respectively. In our implementation, the hidden states serve as the query, while the interaction latents provide both the key and the value.
The operations $\mathrm{SpatialCrossAttn}$, $\mathrm{TemporalCrossAttn}$ and $\mathrm{Mixer}$ are adapted from Stable Video Diffusion ~\cite{blattmann2023stable},

To further refine $\mathbf{V}^\mathrm{latent}$, we utilize hand and face masks to identify the interaction-related regions, which can be formulated as
\begin{equation}
    \mathbf{V}^\mathrm{latent}_\mathrm{m} = \mathbf{V}^\mathrm{latent} \cdot \mathrm{M}_\mathrm{h} \cdot \mathrm{M}_\mathrm{f},
\end{equation}
where $\mathrm{M}_\mathrm{h}$ and $\mathrm{M}_\mathrm{f}$ are extracted from DW-poses~\cite{dwpose} of the driving video. The application of masking is aimed at focusing on the hand and face regions while avoiding disruption to non-interactive areas, thereby accelerating the convergence of the model. Now, we have obtained the region-aware interaction states $\mathbf{V}^\mathrm{latent}_\mathrm{m}$, which captures the spatial and temporal characteristics of the interaction, guiding the denoising process. Finally, the output of the region attention block can be defined as
\begin{equation}
    \label{eq:Fout}
    \mathbf{V}^\mathrm{o}_{att} = \mathbf{V}^\mathrm{i} + \mathbf{V}^\mathrm{latent}_\mathrm{m}.
\end{equation}

\subsection{ID Preserver}
In prior research centered on image-driven generation, the majority of approaches ~\cite{blattmann2023stable, mimicmotion2024,hu2024animateanyone} typically utilize the CLIP image encoder for cross-attention. However, this design is inadequate for effectively maintaining identity consistency in human-centric video generation. To tackle this challenge, we propose an ID preserver to inject ID information into the hidden states. Notably, we employ face embeddings $\mathrm{A}_\mathrm{face}$ extracted via ArcFace ~\cite{deng2019arcface}. The functionality of the ID preserver can be formulated as

\begin{equation}
    \mathrm{A}_\mathrm{emb} = W(\mathrm{A}_\mathrm{face}),
\end{equation}

\begin{equation}
    \mathbf{V}^\mathrm{face} = \mathrm{CrossAttn}(\mathbf{V}^\mathrm{i}, \mathrm{A}_\mathrm{emb}, \mathrm{A}_\mathrm{emb}),
\end{equation}

\begin{equation}
    \mathbf{V}^\mathrm{o}_\mathrm{face} = \mathbf{V}^\mathrm{face} \cdot \mathrm{M}_\mathrm{f},
\end{equation}
% \begin{equation}
% \begin{aligned}
% \mathrm{A}_\mathrm{emb} &= W(\mathrm{A}_\mathrm{face}),\\
% \mathbf{F}^\mathrm{face} &= \mathrm{CrossAttn}(\mathbf{F}^\mathrm{in}, \mathrm{A}_\mathrm{emb}, \mathrm{A}_\mathrm{emb}),\\
% \mathbf{F}^\mathrm{face}_\mathrm{mask} &= \mathbf{F}^\mathrm{face} \cdot \mathrm{M}_\mathrm{face},
% \end{aligned}
% \end{equation}
where $W$ is the projection layers. Notably, $\mathrm{CrossAttn}$ here consists of $\mathrm{SpatialCrossAttn}$, $\mathrm{TemporalCrossAttn}$ and $\mathrm{Mixer}$. The workflow of the ID preserver operates in parallel with the region attention block, and the output from the ID preserver is added to the output of the region attention block. Therefore, the final output as follows:
% we can modify Equation.~\ref{eq:Fout} to obtain $\mathbf{V}^\mathrm{o}$ as follows:
\begin{equation}
    \mathbf{V}^\mathrm{o} = \mathbf{V}^\mathrm{o}_{att} +  \mathbf{V}^\mathrm{o}_\mathrm{face}.
\end{equation}

In this way, the ID preserver effectively utilizes the ID information embedded in the ArcFace representations to impose constraints on the spatial and temporal dimensions of face generation, thereby ensuring facial consistency throughout the animation.

\begin{figure*}[t]
  \centering
  \includegraphics[width=0.9\textwidth]{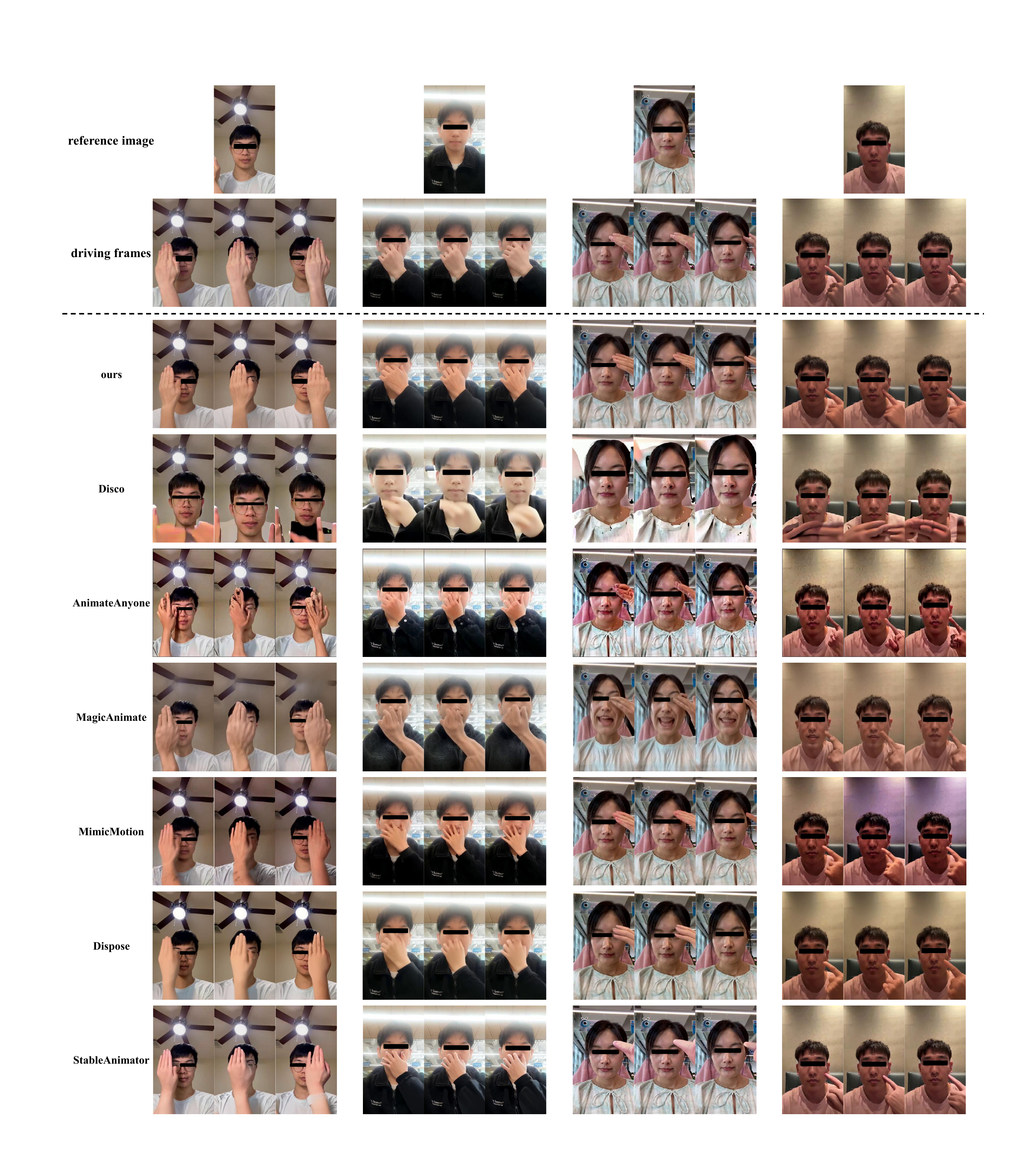}
    \vspace{-10pt}
   \caption{Qualitative comparison vs SOTA methods.}
  \label{fig:comparison}
\end{figure*}

\subsection{Training Objective}

Following the previous methods~\cite{blattmann2023stable,blattmann2023align}, InterAnimate is constructed as a latent video diffusion model. The training objective is formulated by incorporating diffusion loss to guide the denoising process, along with orthogonality loss to maximize the diversity of the patterns learned by the interaction latents.

\textbf{Diffusion Loss.} The diffusion loss is designed to measure the discrepancy between the input video and the reconstructed video in the latent space. Besides, for the interaction animation task, we focus on the performance of the hand and face region. Therefore, we apply loss amplification for these two areas. Specifically, for a given input video $x$, we generate a corresponding latent $z$ using VAE Encoder, then conduct diffusion process by adding noise on $z$. And after multiple denoising steps, we can obtain the denoised video latent $\hat{z}$. The diffusion loss $\mathcal{L}_{\text{diff}}$, which can also be interpreted as a reconstruction loss, is defined as

\begin{equation}
\mathcal{L}_{\text{diff}} = \mathbb{E}_{z}\left[\| z - \hat{z} \|_2^2 \cdot \mathrm{W}_\mathrm{hand} \cdot \mathrm{W}_\mathrm{face} \right],
\end{equation}

\begin{equation}
\mathrm{W}_\mathrm{r} = 
\begin{cases}
\lambda_\mathrm{r} & \text{if}  \  \mathrm{M}_\mathrm{r} > 0, \\
1.0 & \text{ortherwise},
\end{cases}
\end{equation}
where $\|\cdot\|_2^2$, $\mathrm{W}_\mathrm{r}$, $\lambda_\mathrm{r}$ represents the mean squared error, amplification weight and amplification factor, respectively. Specifically, $\lambda_\mathrm{r}$ (\emph{resp.,} $\mathrm{M}_\mathrm{r}$) corresponds to $\lambda_\mathrm{hand}$ (\emph{resp.,} $\mathrm{M}_\mathrm{hand}$) 
and $\lambda_\mathrm{face}$(\emph{resp.,} $\mathrm{M}_\mathrm{face}$), which are applied to amplify the contributions of hand-related and face-related terms, respectively.

\begin{table*}[htbp]
\centering
\captionsetup{font=footnotesize,labelfont=footnotesize,skip=1pt}
\caption{Quantitative comparisons of InterAnimate with the recent SOTA methods.}
\label{tab:quant_comp}
\setlength{\tabcolsep}{8pt}
% \scalebox{1.0}
\resizebox{1.0\linewidth}{!}
{\begin{tabular}{lcccccccc}
\toprule
\multirow{2}{*}{Method}  & \multirow{2}{*}{Publication} & \multicolumn{5}{c}{\textbf{Image}} & \multicolumn{2}{c}{\textbf{Video}} \\ \cmidrule(lr){3-7} \cmidrule(lr){8-9}
 & & \textbf{FID}\ $\downarrow$  & \textbf{SSIM}\ $\uparrow$ & \textbf{PSNR}\ $\uparrow$ & \textbf{LPIPS}\ $\downarrow$ & \textbf{L1}\ $\downarrow$ & \textbf{FID-VID}\ $\downarrow$ & \textbf{FVD}\ $\downarrow$ \\
\midrule
Disco~\citep{Wang_2024_CVPR}  & CVPR 2024 & 116.04 & 0.482 & 18.00 &  0.274 & 3.65E-04 & 126.66 & 1292.37 \\
AnimateAnyone~\citep{hu2024animate} & CVPR 2024 & 112.67 & 0.392 & 13.08 & 0.512 & 2.32E-04 & 100.71 & 964.11\\
MagicAnimate~\citep{xu2023magicanimate} & CVPR 2024 & 45.93 & 0.509 & 15.14 & 0.305 & 1.37E-04  & 33.15 & 446.19 \\ 
MimicMotion~\citep{mimicmotion2024} & arXiv 2024  & 55.02 & 0.458 & 14.85 &  0.363 & 7.50E-05 & 36.87 & 498.77\\
% UniAnimate~\citep{wang2024unianimate} &   &  &  &  &   &  &  & \\
DisPose~\citep{li2025dispose} & ICLR 2025 & 33.30 & 0.565 & 17.04 & 0.246  & 1.07E-04 & 21.22 & 338.40 \\
StableAnimator~\citep{tu2024stableanimator} & CVPR 2025 & 24.95 & 0.748 & 22.24 & 0.162 & 2.69E-05 & 11.80 & 176.20 \\
\midrule
InterAnimate & - & \textbf{23.22} & \textbf{0.785} & \textbf{23.36} & \textbf{0.141} & \textbf{2.27E-05} & \textbf{10.79} & \textbf{122.90}\\
\bottomrule

\end{tabular}
}

\end{table*}

\begin{algorithm}[H]
\caption{Orthogonality Loss Calculation}
\label{alg:orthogonality_loss}
    \renewcommand{\algorithmicrequire}{ \textbf{Input:}}
    \renewcommand{\algorithmicensure}{ \textbf{Output:}} 
    
    \begin{algorithmic}[1]

        \REQUIRE
        $\text{interaction latents } \mathbf{L} \in \mathbb{R}^{n \times d}$
        
        \ENSURE
        $\text{orthogonality loss } \mathcal{L} \in \mathbb{R} $
        
        \STATE $S \gets \mathbf{L} \cdot \mathbf{L}^T \in \mathbb{R}^{n \times n}$ \COMMENT{similarity matrix }
        
        \STATE $I \gets \text{eye}(n) \in \mathbb{R}^{n \times n}$ \COMMENT{identity matrix}
        
        \STATE $mask \gets (1 - \text{eye}(n)) \in \mathbb{R}^{n \times n}$ \COMMENT{mask for non-diagonal}
    
        \STATE $S_{\text{mask}} \gets S[mask]$ \COMMENT{non-diagonal elements of S}
        \STATE $I_{\text{mask}} \gets I[mask]$ \COMMENT{non-diagonal elements of I}

        \STATE $ \mathcal{L} \gets \text{MSE}(S_{\text{mask}}, I_{\text{mask}})$ \COMMENT{Mean Squared Error}
        
        \STATE \textbf{Return: } $\mathcal{L}$
        
    \end{algorithmic}
    
\end{algorithm}

\textbf{Orthogonality Loss.} This loss is introduced to encourage the interaction latents to be orthogonal to each other. This property is particularly important in the context of interaction patterns learning, where the learnable latents should ideally capture diverse and non-redundant interaction information to facilitate high-quality human intertaction animation. The Algorithm. \ref{alg:orthogonality_loss} shows the computation of the orthogonality loss function $\mathcal{F}_{ortho}(.)$. 
% Intuitively, for two interaction latents $\mathbf{L}_i$ and $\mathbf{L}_j$ (where $i \neq j$), the loss is defined as their similarity. The orthogonality loss $\mathcal{L}_{\text{ortho}}$ can be expressed by the following formula: 
% \begin{equation}
% \mathcal{L}_{\text{ortho}} = \mathcal{L}_\mathrm{spa} + \mathcal{L}_\mathrm{temp}.
% \end{equation}
% Algorithm. \ref{alg:orthogonality_loss} details the computation of this orthogonality loss $\mathcal{F}_{ortho}(.)$.
The orthogonality loss is applied separately to two distinct types of latents: spatial latents $\mathbf{L}^\mathrm{s~}$ and temporal latents $\mathbf{L}^\mathrm{t~}$. For each type, the loss is computed to reduce the similarity between any pair of latent vectors within the set. The total orthogonality loss $\mathcal{L}_{\text{ortho}}$ is then the sum of the spatial and temporal components, as given by: 
\begin{equation}
\mathcal{L}_{\text{ortho}} = \mathcal{L}_\mathrm{s} + \mathcal{L}_\mathrm{t},
\end{equation}
where $\mathcal{L}_\mathrm{s}$ (\emph{resp.,} $\mathcal{L}_\mathrm{t}$) calculated from $\mathcal{F}_{ortho}(\mathbf{L}^\mathrm{s~})$ (\emph{resp.,} $\mathcal{F}_{ortho}(\mathbf{L}^\mathrm{t~})$)

This loss encourages the interaction latents to be less correlated, thereby improving the diversity and independence of the learned patterns. Now, we can present the overall objective of our training, which is to minimize the following loss function:
\begin{equation}
\mathcal{L}_{\text{total}} = \mathcal{L}_{\text{diff}} + \beta \cdot  \mathcal{L}_{\text{ortho}},
\end{equation}
where $\beta$ is the weighted factor for the orthogonality loss.

\section{Experiments}

\subsection{Implementation Details}
\textbf{Baselines and test set.} We compare with 6 SOTA human image animation models, including Disco~\citep{Wang_2024_CVPR}, AnimateAnyone~\citep{hu2024animate}, 
MagicAnimate~\citep{xu2023magicanimate}, MimicMotion~\citep{mimicmotion2024},
DisPose~\citep{li2025dispose}  and
StableAnimator~\citep{tu2024stableanimator}. To ensure fairness, we use our InterHF dataset to train the baselines that provide training code. To evaluate the model performance in interaction animation, we construct a test set that includes 18 types of interactions, with 10 videos for each type, sourced from different individuals, for a total of 180 videos. 

\noindent\textbf{Evaluation metrics.} To rigorously evaluate the quality of image generation, we employ frame-level assessment using a suite of standard metrics, including Fréchet Inception Distance (FID)~\citep{NIPS2017_8a1d6947}, Structural Similarity Index Measure (SSIM)~\citep{1284395}, Learned Perceptual Image Patch Similarity (LPIPS)~\citep{Zhang_2018_CVPR}, Peak Signal-to-Noise Ratio (PSNR)~\citep{5596999}, and L1 distance. For video evaluation, we construct clips by concatenating every 16 consecutive frames, and compute Video Fréchet Inception Distance (FID-VID)~\citep{ijcai2019p276} and Fréchet Video Distance (FVD)~\citep{unterthiner2018towards} to jointly assess visual fidelity and temporal coherence. We utilize the evaluation code from Disco ~\citep{Wang_2024_CVPR}.

% \noindent\textbf{Dataset.} We trained the models on the InterHF training set for interaction aniamtion. The training set consists of about 6K videos that capture a single person performing hand-face interaction. To evaluate the model performance in interaction animation, we construct a test set that includes 18 types of interactions, with 10 videos for each type, sourced from different individuals, for a total of 180 videos. 

\noindent\textbf{Training.} We choose the Stable Video Diffusion~\cite{blattmann2023stable} as our architecture, and adopt the pre-trained weights from the public stable video diffusion 1.1 image-to-video model. Besides, we leverage the sampling strategy of Mimicmotion ~\cite{mimicmotion2024}. The training is an end-to-end process, and the trainable components include denoising U-Net, pose encoder, learnable interaction latents and ID preserver. We train our model on 8 NVIDIA A100 GPUs for 50K steps with image size 384 $\times$ 640, learning rate 1e-5, and a per-device batch size of 1. During training, a random frame from the video is selected as the reference, and a 16-frame clip is sampled as the target.  In our experiments, the learnable interaction latents have a dimension of $d = 512$, while both spatial and temporal latents consist of 512 vectors each. During the soft quantization process, we employ a temperature parameter $\tau = 1.0$. We set $\alpha$ in $ \mathrm{Mixer}$ to 0.5 in our experiments. The loss amplification factors are set to $\lambda_{\mathrm{hand}} = 5.0$ and $\lambda_{\mathrm{face}} = 2.0$, respectively. Additionally, the orthogonality loss factor is established at $\beta = 0.0001$.

\vspace{-0.2cm}
\subsection{Qualitative Comparison}

As illustrated in Fig. \ref{fig:comparison}, we present 4 hand-face interaction actions from left to right, including the palm moving in front of the face, pinching the nose, touching the eyebrow, and poking the face. The comparison focuses on each method's ability to handle facial consistency, hand structure, and interaction quality. 

It is clear that the human identity in our animation results maintains a high level of consistency with the reference images. Moreover, the generated hand structure is reasonable and natural. Impressively, our method excels at producing high-quality hand-face interactions. For example, in the challenging task of pinching the nose, our method not only maintains accurate hand geometry but also realistically captures the squeezing and releasing motion. Additionally, in the example of poking the face, our method produces lifelike facial deformations, effectively simulating the compression of facial muscles. In contrast, other methods often perform inconsistently in interaction animation.

Disco keeps facial visibility but fails to generate complete hands, and avoids hand-face contact, limiting expressive interaction. AnimateAnyone follows driving actions and maintains facial identity, but produces implausible hand structures. Magicanimate aligns with interaction action but suffers from unnatural hands and distorted facial appearance. MimicMotion improves identity consistency and hand quality, yet lacks fluidity and realism in hand-face contact. Similarly, Dispose and StableAnimator enhance hand quality, but often produce stiff, unnatural interaction.

% Disco keeps facial non-occlusion, and exhibits a reluctance to perform hand-face interactions, limiting the expressiveness of interaction animations. In addition, Disco cannot produce the complete hands.  AnimateAnyone effectively follows driving interactions and maintains good facial identity consistency, but fails to produce plausible hand structures, resulting in noticeable artifacts that detract from the overall interaction quality. Magicanimate aligns with interaction action but exhibits unnatural hand structures and poor facial consistency, leading to significant variations in identity that undermine coherence in interaction animations. MimicMotion shows improvements in ID consistency and hand quality. however, it still lacks fluidity and realism in hand-face interactions, which diminishes the naturalness of the animation. Similarly, both Dispose and StableAnimator further enhance hand quality and interaction quality yet often produce movements that appear forced, impacting the overall realism of the interaction animations. In contrast, our method excels across all metrics by maintaining facial consistency, generating realistic hand structures, and facilitating fluid hand-face interactions, ultimately delivering coherent, high-quality expressive interaction animation.

\subsection{Quantitative Comparison}

The quantitative results are summarized in Table \ref{tab:quant_comp}, where multiple metrics are reported for both image and video quality.
In terms of image quality metrics, InterAnimate outperforms all other methods. InterAnimate achieves the lowest FID (23.22), indicating high visual fidelity and alignment with real data. InterAnimate also records the highest SSIM (0.785) and PSNR (23.36), reflecting strong preservation of fine details and structure, which is crucial for realistic hand-face interactions like finger articulation and facial muscle movement. Additionally, the lowest L1 error and LPIPS further highlight InterAnimate’s precision and minimal distortion in reconstructing complex interactions.
% In terms of image quality metrics, InterAnimate achieves superior performance across all results compared to existing methods. Specifically, it achieves the lowest FID score of 23.22, indicating exceptional visual fidelity and alignment with real data distributions. This result underscores the superior realism and faithfulness of our method in generating hand-face interaction animations. Additionally, InterAnimate achieves the highest SSIM of 0.785 and PSNR of 23.36, highlighting its exceptional ability to preserve fine details and structural integrity in generated images. These metrics are particularly crucial for capturing subtle nuances, such as finger articulations and facial muscle movements, which are essential for realistic hand-face interaction. Furthermore, InterAnimate demonstrates the smallest L1 error and LPIPS, further emphasizing its precision in reconstructing intricate interactions with minimal distortion.
For video quality assessment, InterAnimate achieves the lowest FID-VID (10.79) and FVD (122.90), demonstrating its strength in generating temporally coherent and realistic videos. These results highlight its effectiveness in capturing natural motion in hand-face interactions, such as the smooth transition of fingers touching or moving away from the face.

Notably, while StableAnimator and DisPose perform competitively on certain metrics, they fall short in balancing both static image quality and dynamic video coherence. This trade-off is especially limiting for interactions, where fine detail and smooth motion transitions are essential. In contrast, InterAnimate’s strong performance across all metrics highlights its ability to generate high-fidelity, temporally consistent animations, making it well-suited for applications requiring intricate hand-face interactions.
% Notably, while methods like StableAnimator and DisPose perform competitively on certain metrics, they fall short in balancing both static image quality and dynamic video coherence. These improvements are particularly significant for hand-face interaction, where precise detail and smooth motion transitions are critical. Overall, the superior performance of InterAnimate across all metrics underscores its effectiveness in addressing the challenges of high-fidelity and temporally consistent video generation, making it a compelling choice for applications requiring intricate hand-face interactions.

\subsection{Ablation Study}

The ablation study in Table \ref{tab:ablation} examines the impact of key components, including the Region-aware Interaction System (RIS), soft quantization (quantize), orthogonality loss (o-loss), and ID Preserver (ID).
The results demonstrate that RIS is essential for interaction animation, as its removal leads to a substantial decline across all image and video metrics. This highlights its importance in capturing fine-grained interaction details and maintaining temporal coherence in hand-face interactions.
Removing quantization from the region attention block also causes a sharp decline in performance, highlighting its indispensable role in the extraction of priors from interaction latents.
After removing the orthogonality loss and ID Preserver, there is a slight decrease in some metrics, with improvements observed in L1 and FID-FVD. However, the FVD value increased significantly (from 122.90 to 174.02 and 165.03, respectively), indicating that without o-loss, the interaction latents fail to capture sufficient interactive information, resulting in degraded performance over time. Similarly, temporal ID inconsistency also leads to a degradation in the FVD metric.

% After removing the orthogonality loss and ID preserver, there is a slight decrease in metrics, and even improvements were observed in L1 and FID-FVD. However, we notice that the FVD value increased significantly (from 122.90 to 174.02 and 165.03, respectively). This indicates that without the orthogonality loss, the interaction latents cannot learn enough diverse interactive information, leading to poor performance over time. Additionally, after removing the ID preserver, the ID consistency over time also worsened.

\begin{table}[t]
\centering
\captionsetup{font=footnotesize,labelfont=footnotesize,skip=1pt}
\caption{Ablation analysis of the key components.}
\label{tab:ablation}
% \setlength{\tabcolsep}{8pt}
% \scalebox{1.0}
\resizebox{1.0\linewidth}{!}
{\begin{tabular}{lccccccc}
\toprule
\multirow{2}{*}{} & \multicolumn{5}{c}{\textbf{Image}} & \multicolumn{2}{c}{\textbf{Video}} \\ \cmidrule(lr){2-6} \cmidrule(lr){7-8}
 & \textbf{FID}$\downarrow$ & \textbf{SSIM}$\uparrow$ & \textbf{PSNR}$\uparrow$ & \textbf{LPIPS}$\downarrow$ & \textbf{L1}$\downarrow$ & \textbf{FID-VID}$\downarrow$ & \textbf{FVD}$\downarrow$ \\
\midrule
w/o RIS & 31.25 &  0.709 & 20.53 & 0.254 & 2.68E-05 & 17.48 & 242.31\\
w/o quantize& 29.66 & 0.745 & 21.84 & 0.168 & 2.52E-05 & 12.88 & 175.13\\
w/o o-loss & 25.76 & 0.772 &  23.29 & 0.153 & \textbf{2.01E-05} & 10.34 & 174.02\\
w/o ID & 25.77 & 0.767 & 23.05 & 0.153  & 2.11E-05 & \textbf{10.31} & 165.03\\
ours & \textbf{23.22} & \textbf{0.785} & \textbf{23.36} & \textbf{0.141} & 2.27E-05 & 10.79 & \textbf{122.90}\\
\bottomrule
\end{tabular}}

\end{table}

\begin{figure}[H]
    \centering
    \includegraphics[width=\linewidth]{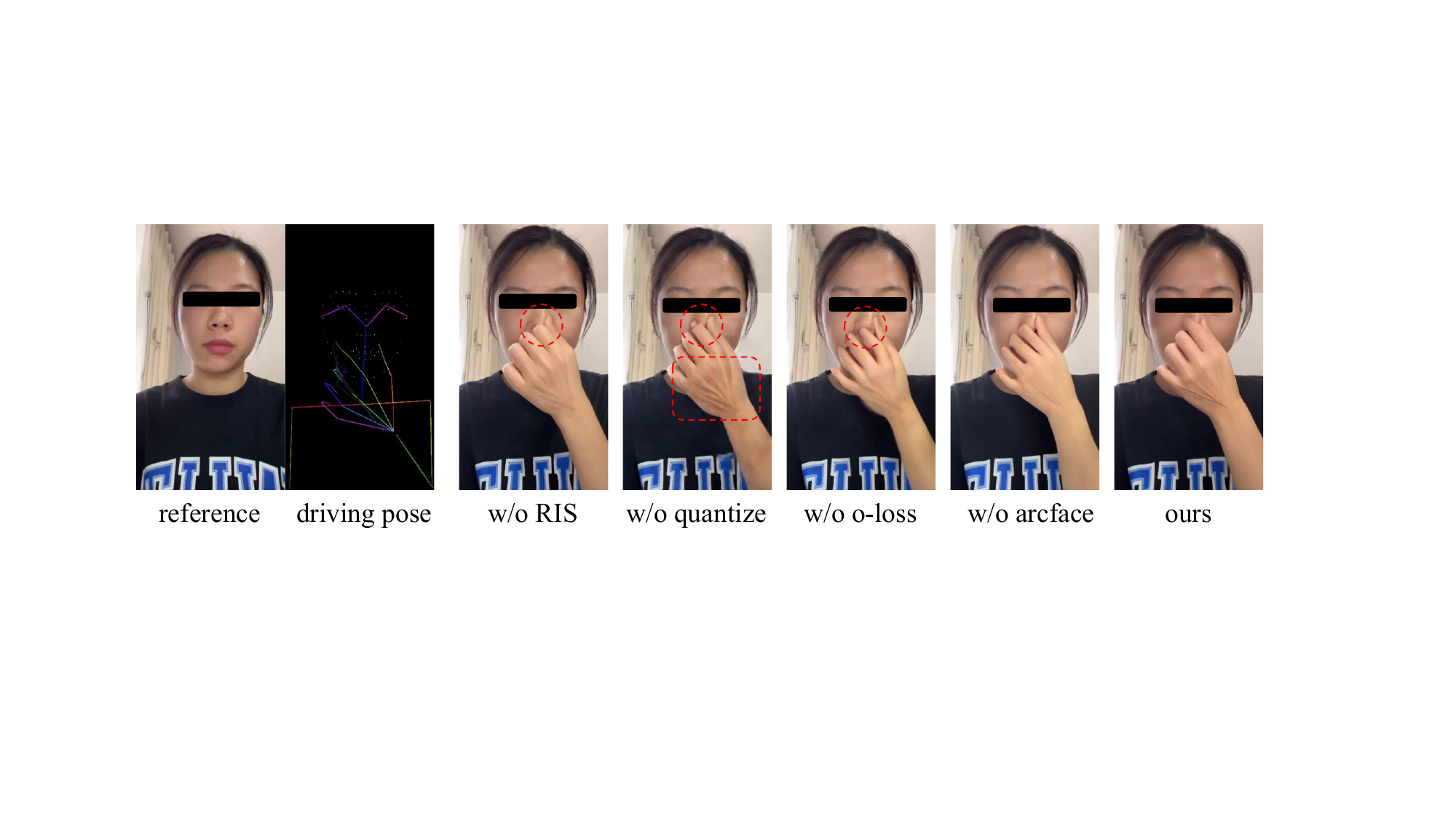}
    \caption{The visualization of the ablation.}
    \label{fig:ablation}
    \vspace{-0.2cm}
\end{figure}

As illustrated in Fig. \ref{fig:ablation}, without RIS, the hand structure remains reasonable, but the fingers fail to pinch the nose. This indicates that while the model can learn hand priors from the interaction dataset, it lacks effective interaction modeling.
Removing quantization results in nose distortion and degraded hand structure, suggesting the model struggles to exploit priors from interaction latents. Similar distortions are observed when orthogonality loss is removed.

\section{Conclusion and Discussion}

In this paper, we address a critical gap in human video generation by focusing on the interactions between hands and faces. We introduce InterHF, a comprehensive dataset that captures diverse hand-face interaction patterns, and present InterAnimate, a region-aware diffusion model designed specifically for interaction animation. Our extensive evaluations demonstrate the effectiveness of our approach, showcasing its potential for creating more realistic and natural animations. This work sets the stage for future research into complex interactions in human video generation.

To improve the interaction animation further, we will make two attempts in the future. Firstly, we intend to expand our interaction video dataset with a wider variety of interaction types and more intricate interaction modes. Secondly, we will develop models that are stronger in perception of interactions.

\nocite{ma2024follow, zeng2023face, han2024face, li2024uv, guo2024liveportrait, ouyang2024mergeup, li2023finedance, li2024lodge, li2024lodge++, li2024interdance, zhou2024uniqa, zhou2023etdnet, zhou2024unihead, xue2024follow, ma2024followyourclick, wang2024cove, wang2024taming, lin2024consistent123, lin2025mvportrait, lin2023rich, lyu2025hvis}

\clearpage

\begin{acks}
This work was supported by Ant Group Research Intern Program and the Pioneer R\&D Program of Zhejiang Province(No. 2024C01024). This work was also partially  supported by Shenzhen Key Laboratory of next generation interactive media innovative technology(No: ZDSYS20210623092001004).
\end{acks}

\bibliographystyle{ACM-Reference-Format}
\balance
\bibliography{reference}

% \appendix

\end{document}